\documentclass[times, review, 10pt]{elsarticle}
\usepackage{setspace}
\usepackage{hyperref}
\usepackage{amsmath}
\usepackage{array}
\usepackage{graphicx}
\usepackage{epstopdf}
\usepackage{amssymb}
\usepackage{float}
\usepackage{enumerate}
\usepackage{booktabs}
\usepackage{adjustbox}
\usepackage{enumitem}
\usepackage{balance}
\usepackage{xr}
\usepackage{subcaption}
\usepackage{booktabs}
\usepackage{adjustbox}
\usepackage{multirow}
\usepackage{multicol}
\usepackage{threeparttable}
\usepackage{orcidlink}
\usepackage{caption}
\usepackage{pifont}
\biboptions{sort&compress}
\journal{LATEX} 

\hypersetup{
	colorlinks=true,
	linkcolor=black,
}

\begin{document}
\doublespacing

\begin{frontmatter}

\title{XOResNet: Exclusive-OR Meta-Residuals Facilitate Deep Spiking Neural Networks Learning}


\author[mainaddress,secondaryaddress]{Jianfang~Wu$^{\orcidlink{0000-0001-7213-5877}}$}
\author[mainaddress,secondaryaddress]{Junsong~Wang$^{\orcidlink{0000-0002-4846-6585}}$\corref{correspondingauthor}}
\cortext[correspondingauthor]{Corresponding author}
\ead{wangjunsong@sztu.edu.cn}
\address[mainaddress]{School of Artificial Intelligence, Shenzhen Technology University, Shenzhen 518118, China}
\address[secondaryaddress]{Faculty of Data Science, City University of Macau, Macau 999078, China}


\begin{abstract}	
Spiking neural networks (SNNs) hold promise for demonstrating superior learning and representation capabilities in deep models. Given the tremendous success of ResNet in deep learning, it would naturally follow to train deep SNNs with residual learning. However, existing residual structures for constructing deep SNNs still present challenges of spike redundancy or information loss, as well as redundant learning. In the present study, we first aim to address issues of relative spike redundancy in identity mapping and information loss in non-identity mapping. To this end, we propose an OR-ADD (OA) shortcut connection to merge output spikes/currents from two branches in the residual structure. Furthermore, to mitigate redundant learning in the backbone branch of the residual structure, we introduce the concept of XOR meta-residuals, i.e., selecting pre-learning residuals using the Exclusive-OR (XOR) operation for the backbone branch. Finally, by integrating the OA shortcut and XOR meta-residuals, we devise the XOR residual block and further construct XOResNet with varying depths based on this block. Extensive experiments on four datasets, Fashion-MNIST, CIFAR-10, CIFAR-100, and miniImageNet, show that the proposed XOResNet outperforms existing state-of-the-art deep SNNs optimized via gradient descent. These results validate the effectiveness of our OA shortcut and XOR meta-residual components in overcoming fundamental limitations of residual learning in SNNs, providing new architectural insights for building high-performance neuromorphic systems. 
\end{abstract}


\end{frontmatter}


\section{Introduction}
\label{introduction}
Inspired by the working mechanisms of the human brain and the working patterns of biological neurons, spiking neural networks (SNNs) are considered a promising model in artificial intelligence, embodying high efficiency akin to the brain~\cite{1}. Meanwhile, SNNs are also considered the third generation of neural networks due to their energy advantage of asynchronous binary spiking communication and powerful representation of spatio-temporal dynamics~\cite{2}. By drawing on and mimicking the learning algorithms and network structure of artificial neural networks (ANNs), SNNs exhibit performance close to that of ANNs on some classification tasks but are still inferior to ANNs on complex tasks~\cite{3,4,6, 7,8}. An important reason is that discrete spikes and complex spatio-temporal dynamics limit SNNs from directly adopting the deep construction method of ANNs. However, deep networks have advantages over shallow networks in terms of computational cost and representational capability~\cite{5}.

Artificial neural networks have achieved great success in various tasks, largely due to the success of deep learning. The depth of a network is closely related to its performance on practical tasks, while the function represented by a deep network requires a single hidden-layer network constructed from an exponential number of neural units to be comparable~\cite{12}. To solve the gradient problem in deep neural networks, He \textit{et al.} \cite{13} proposed the concept of residual learning and used the residual structure to construct ``very deep" networks. Therefore, residual structure is widely used in the construction of deep neural networks and catalyzes the rapid development of deep learning.

To achieve higher performance in SNNs, it will be natural to construct deeper networks with residual structure. The spiking version of ResNet (Spiking ResNet)~\cite{15,16,18} achieves state-of-the-art performance on most datasets by replacing nonlinear activation units in ANNs with spiking neurons. However, Spiking ResNet still suffers from performance degradation caused by gradient issues. Meanwhile, simply transplanting shortcut connections from ANNs can disrupt the spike-based identity mapping. To train deep SNNs using spatio-temporal backpropagation (STBP)~\cite{23}, Fang \textit{et al.} \cite{4} solved the gradient problem by summing output spikes from two branches, constructing SEW ResNet, a model of deep SNNs with over 100 layers. However, this spike-summing approach can lead to non-spike computation, harming deployment on neuromorphic chips that process binary inputs~\cite{21}. To preserve the binary property of spikes in shortcut connections, Shan \textit{et al.} \cite{22} used the OR operation to merge output spikes from two branches. This maintains the binary property of spikes while reducing spike redundancy. However, when the shortcut connection involves non-identity mappings with scaling transformations, the OR operation may result in the loss of joint information from both branches.

Furthermore, existing methods for constructing deep SNNs have almost entirely adopted the residual structure of ResNet~\cite{13}. This may result in spike redundancy in the residual learning of the backbone branch relative to the shortcut branch. The residual branch (backbone branch) in ResNet specializes in capturing residual features - specifically, the differential components between the layer input and the identity shortcut output. However, only after the information from the two branches is fused can the learned residual information be truly determined. In other words, the backbone branch does not determine in advance what residual information needs to be learned, i.e., the residuals of the backbone branch are post-learning residuals. In the case of information transmission via binary spikes, there may be relative redundancy of spikes in the two branches. This may also result in redundant learning in the backbone branch. To reduce spike redundancy and redundant learning, the backbone branch should be provided with pre-learning residual guidance, i.e., selecting pre-learning residuals.

In the present study, we consider two cases of shortcut connection in the construction of deep SNNs: spike operation in identity mapping and information retention and utilization in non-identity mapping. To reduce spike redundancy of the backbone branch relative to the shortcut branch and facilitate residual learning of the backbone branch, we propose utilizing the Exclusive-OR (XOR) operation to provide pre-learning residuals, i.e., meta-residuals, for the backbone branch. We construct residual blocks using the aforementioned method and use them to construct deep SNNs called XOResNet, which consistently outperform both OA ResNet and OR ResNet. We deepen XOResNet to 110 layers without encountering any degradation problem, and theoretically, it can be deepened to any desired depth.

The main contributions and highlights of this study can be summarized as follows:
\begin{enumerate}[label=(\roman*)]
	\setlength\itemsep{0.25em}
	\item For the shortcut connection of residual structures, we propose the OR-ADD (OA) connection method. If the shortcut branch realizes spike identity mapping, the output spikes of the shortcut and backbone branches are merged by the OR operation, achieving information complementation while maintaining the binary property of spikes. If the shortcut branch realizes non-identity mapping with scale transformation, the sum of output currents from both branches is used as input to the spiking neuron to avoid information loss.
	\item For the residual learning in the backbone branch, we propose to utilize the XOR operation to pre-screen the residual features that require learning, thereby providing pre-learning residuals ((i.e., meta-residuals)) for the backbone branch. This approach aims to reduce redundant learning in the backbone branch and enhance its residual learning capabilities.
	\item We integrate the OA shortcut and XOR meta-residuals to construct deep SNNs called XOResNet. Extensive comparisons across four benchmark datasets (CIFAR-10, CIFAR-100, Fashion-MNIST, and miniImageNet) reveal that XOResNet consistently outperforms both OA ResNet and OR ResNet. This demonstrates the efficiency of our proposed residual structure.
\end{enumerate}

The remainder of this paper is organized as follows. Section~\ref{sec:Related Work} is an overview of related work on building deep SNNs. In Section~\ref{sec:Methods}, we describe in detail the proposed shortcut branch connection method OR-ADD(OA), the residual information extraction method, and the constructed XOResNet network. In Section~\ref{sec:Experiments and results}, we systematically present the datasets and experimental results. In Section~\ref{sec:discussion}, we present a detailed discussion of our work. Finally, in Section~\ref{sec:conclusion}, we present our conclusions and further work.

\section{Related Works}
\label{sec:Related Work}

Deep networks offer advantages over shallow networks in terms of computational cost and representation ability. Research efforts in constructing deep SNNs can be categorized into two main classes: (1) converting pre-trained deep ANNs into SNNs, and (2) training deep SNNs with residual structures by spatio-temporal backpropagation (STBP).


\textbf{ANN to SNN conversion (ANN2SNN)} \quad ANN2SNN replaces the non-linear activation units of the pre-trained source ANN with spiking neurons~\cite{32,33}. The central idea of this approach is to use the firing rate of spiking neurons or the average postsynaptic potential to approximate ReLU activation in artificial neurons~\cite{34}. Some advanced conversion works on VGG and ResNet architectures with near-lossless accuracy by adding scaling tricks like weight normalization and threshold balancing~\cite{15,16,18,35}. However, hundreds or thousands of firing statistics must be performed on spiking neurons to approximate their firing rate to the activation output of a ReLU.

\textbf{Training deep SNNs with residual structures based on STBP} \quad The nondifferentiable binary spike activity leads to the inability to train SNNs directly by backpropagation (BP)~\cite{36, 37} algorithms, however, the great success of BP in training deep ANNs is very tempting for the training of SNNs. An algorithm called spatio-temporal backpropagation~\cite{23} was proposed for training SNNs by introducing the gradient of a differentiable function to surrogate the gradient of the Heaviside step function in the error backpropagation process. Among works on training deep SNNs based on STBP, Spiking ResNet~\cite{15,16,18}, a spiking version of ResNet that fully adopted the connection structure of ResNet, still suffered from performance degradation caused by the gradient problem. The SEW ResNet solves the gradient issue in Spiking ResNet by summing output spikes from the backbone and shortcut branches, enabling depths beyond 100 layers~\cite{4}. However, spike summation destroys the binary nature of spikes ($1+1=2$), hindering deployment on neuromorphic chips with binary inputs~\cite{21}. To preserve the binary nature of spikes, Shan \textit{et al.} \cite{22} proposed using an OR operation to merge the output spikes from the backbone branch and the shortcut branch. However, when the shortcut connections are non-identity mappings with scale transformations, the OR operation may result in joint information loss from both branches. Hu \textit{et al.} \cite{38} directly merged the output currents of the two branches, proposing MS-ResNet, which avoids some problems caused by spike operations. Subsequently, this residual connection approach was also used to construct the spiking version of Transformer~\cite{39}. However, this may also result in redundancy of output spikes during identity mapping. Furthermore, the aforementioned works do not account for the specificities and advantages of spiking communications, resulting in a certain degree of redundant learning in the backbone branch.

\section{Methods}
\label{sec:Methods}
 \subsection{Shortcut connections in residual structures}
 \label{OR-ADD}
\begin{figure}[!ht]
	\centerline{\includegraphics[width=0.5\textwidth]{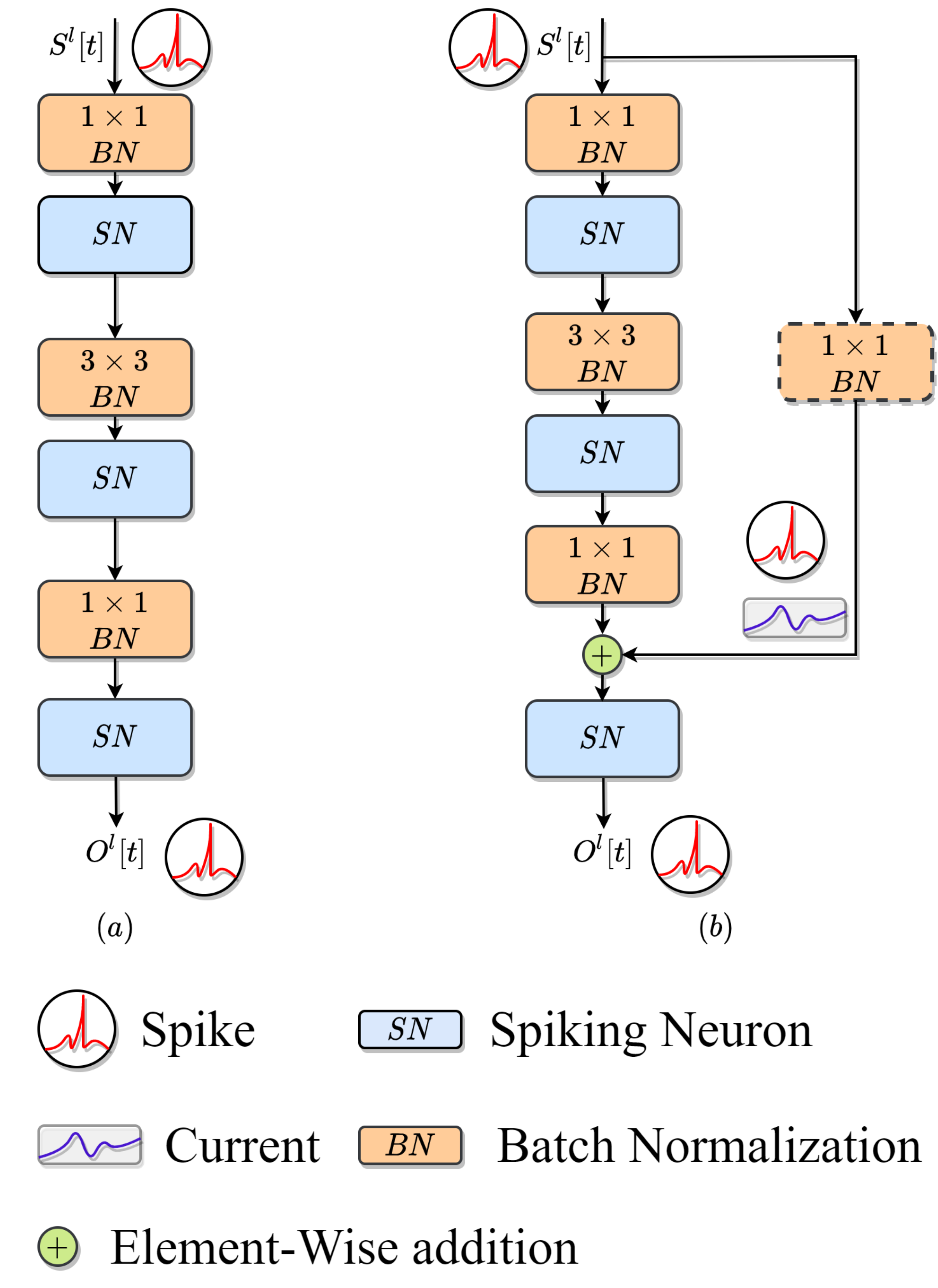}}
	\caption{The basic building blocks for deep SNNs. (a) The basic building block in Plain Network. (b) The basic building block in Spiking ResNet. $S^{l}[t]$/$O^{l}[t]$ denotes the input/output spikes of layer $l$ at time $t$. $1\times1$ and $3\times3$ denote the convolution kernel size. $BN$ is a batch normalization operation. $SN$ denotes the spiking neuron.}
	\label{fig1}
\end{figure}

\begin{figure}[!ht]
	\centering
	\begin{subfigure}[b]{0.45\textwidth}
		\includegraphics[width=\textwidth]{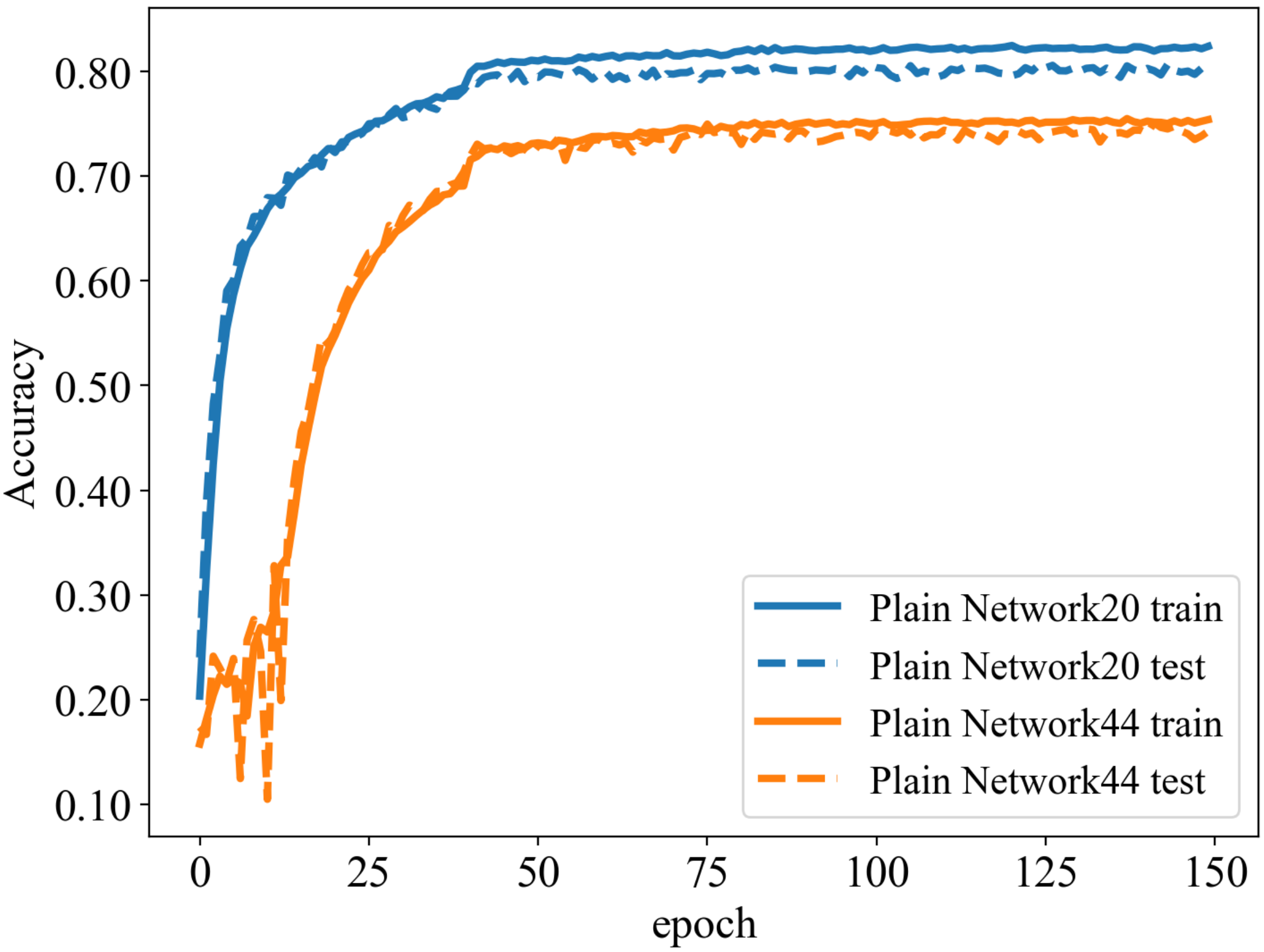}
		\caption{Plain Network}
	\end{subfigure}
	\hspace{0.3cm}
	\begin{subfigure}[b]{0.45\textwidth}
		\includegraphics[width=\textwidth]{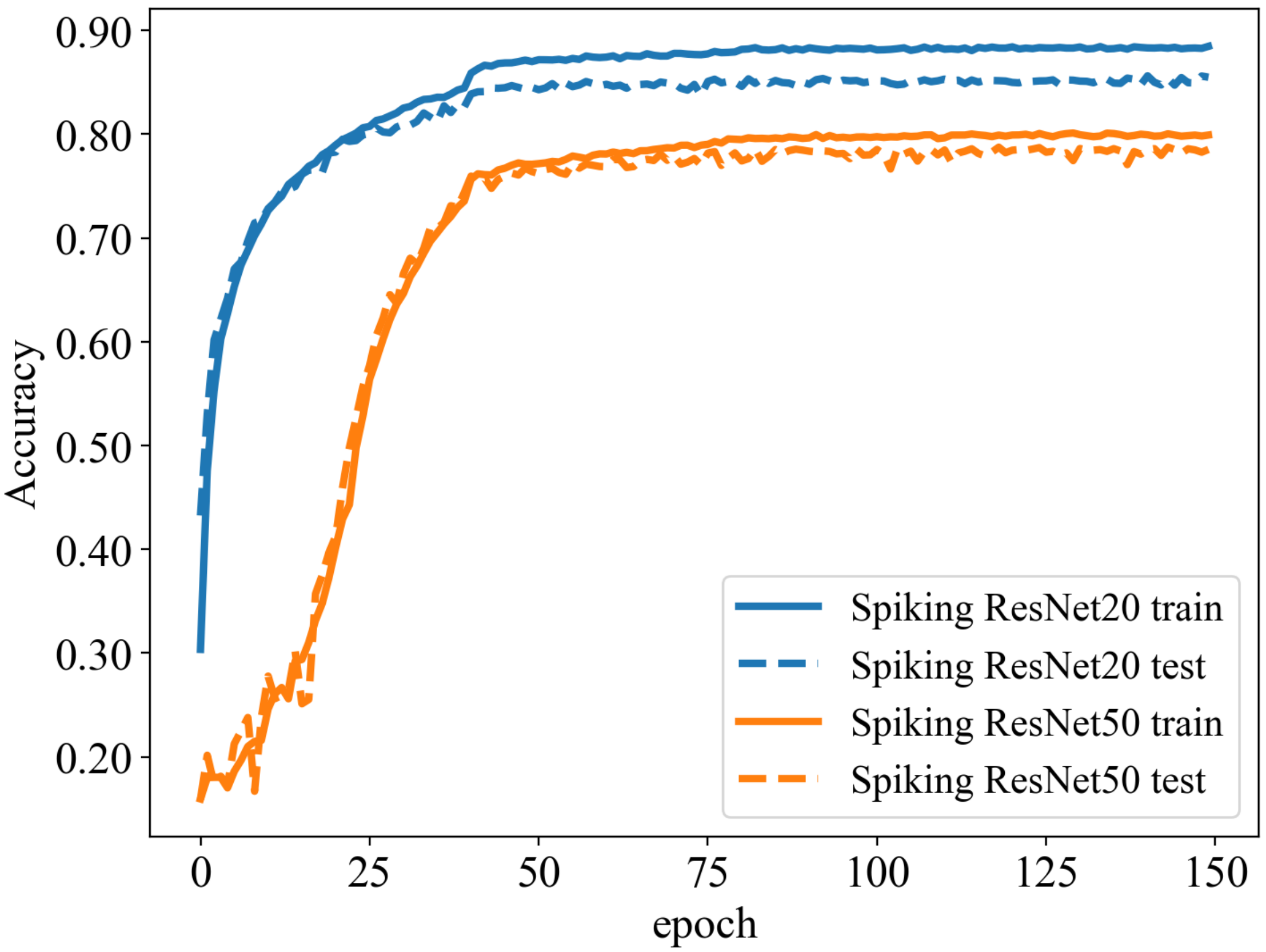}
		\caption{Spiking ResNet}
	\end{subfigure}
	\caption{Training accuracy and test accuracy of models with different depths on CIFAR-10.}
	\label{fig2}
\end{figure}

 The fact that deep neural networks have demonstrated more powerful representation ability than shallow networks also holds for SNNs. Simply increasing the depth of SNNs inevitably suffers from the performance degradation problem experienced by ANNs. Ignoring the differences between SNNs and ANNs and duplicating ResNet's residual structure exactly also fails to solve the performance degradation problem of deep SNNs in gradient-based training. The networks constructed from the two basic building blocks are Plain Network and Spiking ResNet, as shown in Fig.~\ref{fig1}, but both deep networks suffer from performance degradation (shown in Fig.~\ref{fig2}).

In this work, we consider the specificity of spike communication in SNNs. For the shortcut connection in a residual structure, we propose the OR-ADD (OA) connection method. Specifically, when spikes realize identity mapping (im), we merge the output spikes of the shortcut branch and backbone branch via the OR operation, maintaining the binary property of spikes and avoiding spike redundancy. When the shortcut connection is a non-identity mapping (nim) with scale transformation, the output currents from the two branches are summed as input to the spiking neuron (SN), avoiding information loss.

\begin{figure}[!ht]
	\centerline{\includegraphics[width=\textwidth]{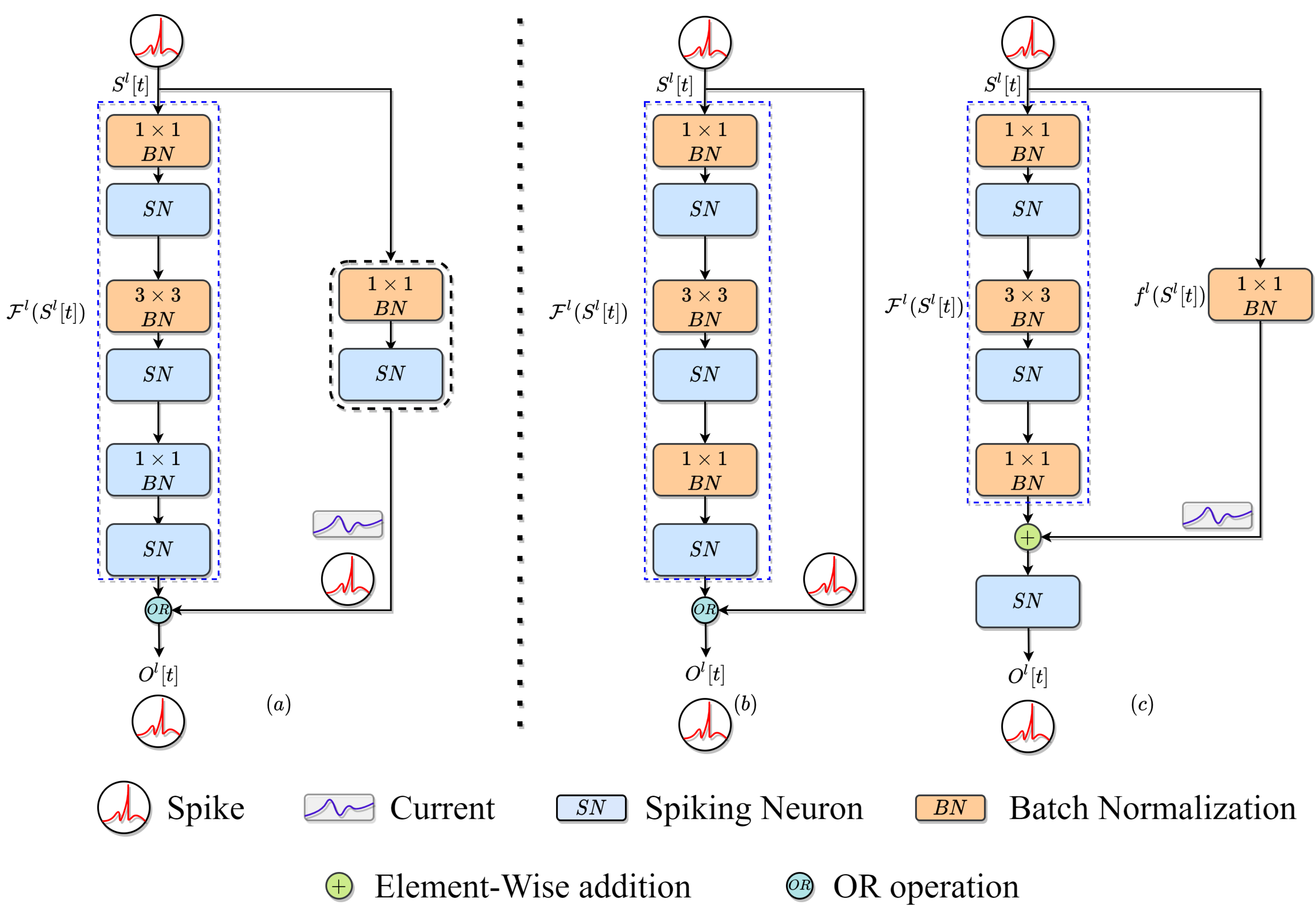}}
	\caption{Shortcut connection. (a) OR shortcut connection. (b) (c) OR-ADD (OA) shortcut connection.}
	\label{fig3}
\end{figure}

\begin{table}[!hb]
	\renewcommand{\arraystretch}{0.8}
	\centering
	\captionsetup{skip=2pt}
	\caption{Interconversion between logical operations and arithmetic operations.}
	\label{tab1}
	\resizebox{0.7\textwidth}{!}
	{
		\begin{threeparttable}
			\begin{tabular}{ccc}
				\toprule
				\textbf{Name} & \textbf{Logical expression}&\textbf{Arithmetic expression}\\
				\midrule
				OR&A$\vee$B&A+B-AB\\
				XOR&A$\oplus$B&A+B-2AB\\
				\bottomrule
			\end{tabular}
		\end{threeparttable}
	}
\end{table}

As shown in Fig.~\ref{fig3}(a), the OR shortcut connection proposed in \cite{22} is effective for merging output spikes and maintaining the binary property when the shortcut realizes identity mapping. However, for a non-identity mapping shortcut, this may result in information loss. Combining the dynamical properties of SN, logical operations (Table~\ref{tab1}), and our proposed OA connection method (Fig.~\ref{fig3}(b) and Fig.~\ref{fig3}(c)), for non-identity mapping:
\begin{equation}
	\label{equ1}
	\begin{aligned}
		SN(f^{l}(S^{l}[t])) &\ne f^{l}(S^{l}[t])\ne S^{l}[t],\\
		SN(\mathcal{F}^{l}(S^{l}[t])) \vee 	SN(f^{l}(S^{l}[t]))  &\ne  SN(\mathcal{F}^{l}(S^{l}[t])+f^{l}(S^{l}[t])) . 
	\end{aligned}
\end{equation}
$SN(\cdot)$ is a non-linear activation unit that produces spiking outputs of 0 or 1. Therefore, we propose the OA shortcut connection method, for which different spike/current merging methods are used for different shortcut connection purposes. As shown in Fig.~\ref{fig3}(b) and Fig.~\ref{fig3}(c), the OA connection method can be expressed as follows:
\begin{equation}
	\label{equ2}
	OA(S^{l}[t],\mathcal{F}^{l}(S^{l}[t]))=
	\begin{cases}
	\mathcal{F}^{l}(S^{l}[t]) \vee S^{l}[t]= \mathcal{F}^{l}(S^{l}[t]) + S^{l}[t] - \mathcal{F}^{l}(S^{l}[t]) \cdot S^{l}[t]\quad \text{im} \\
	\qquad \qquad \qquad SN(\mathcal{F}^{l}(S^{l}[t])+f^{l}(S^{l}[t]))  \quad \text{nim}
	\end{cases}.
\end{equation}

\subsection{The information retention advantages of OA connections}
\label{advantages}

Next, from an information preservation perspective, we theoretically demonstrate that the ADD operation outperforms the OR operation in shortcut connections involving non-identity mappings with scale transformations. In Fig.~\ref{fig3}(c), for a non-identity mapping with scale transformation, the joint distribution of the inputs 
$\mathcal{F}^{l}(S^{l}[t])$ and $f^{l}(S^{l}[t])$ is given by $p(\mathcal{F}^{l}(S^{l}[t]),f^{l}(S^{l}[t]))$. For simplicity, we denote $\mathcal{F}^{l}(S^{l}[t])$ by the random variable $X_1$  and $f^{l}(S^{l}[t])$ by $X_2$.  If the OR operation is applied to merge the spiking outputs of the backbone branch and the shortcut branch, the firing probability of   $O^l_{OR}[t]$ is:
\begin{equation}
	\label{1}
	 p\left(O^l_{OR}[t] =1\mid X_1,X_2 \right)=p\left(X_1\right)+p(X_2)-p(X_1)\cdot p(X_2) .
\end{equation}
The conditional entropy is:
\begin{equation}
	\label{2}
	\begin{aligned}
		H\left(O^l_{OR}[t] \mid X_1,X_2 \right)&=H_B\left(p(X_1)+p(X_2)-p(X_1)\cdot p(X_2) \right),\\
		H_{_B}\left(p\right)&=-p\log p-(1-p)\log(1-p).
	\end{aligned}
\end{equation}
Since the output $O^l_{OR}[t]$ depends solely on the logical OR operation of $SN(X_1)$ and $SN(X_2)$, it cannot distinguish whether the spiking event is triggered by $X_1$ or $X_2$, leading to a loss of joint information. If the ADD operation is applied to sum the currents, the merged current is then fed as input to $SN(\cdot)$. The firing probability of $ O^l_{ADD}[t]$ can then be expressed as:
\begin{equation}
	\label{3}
	P\left(O^l_{ADD}[t]=1|X_{1},X_{2}\right)=p\left(X_{1}+X_{2}\right).
\end{equation}
The conditional entropy is:
\begin{equation}
	\label{4}
	H\left(O^l_{ADD}[t]\mid X_1,X_2\right)=H_B\left(p\left(X_1+X_2\right)\right).
\end{equation}
The output $O^l_{ADD}[t]$ directly reflects the summation of inputs, preserving the linear combination information from $X_1$ and $X_2$. Even when individual inputs $X_1$ or $X_2$ are subthreshold (insufficient to trigger a spike), their superposition $X_1 + X_2$ may still evoke a firing event, thereby capturing joint information.

According to Jensen's inequality:
\begin{equation}
	\label{4}
	p\left(X_1\right)+p\left(X_2\right)-p\left(X_1\right)\cdot p\left(X_2\right) \leq p\left(X_1+X_2\right).
\end{equation}
This demonstrates that output $O^l_{ADD}[t]$ exhibits a higher firing probability, with the information entropy satisfying $H(O^l_{ADD}[t]) \ge H(O^l_{OR}[t])$, while the conditional entropy shows $H\left(O^l_{ADD}[t]\mid X_1,X_2\right) \leq H\left(O^l_{OR}[t] \mid X_1,X_2 \right)$. The mutual information can be computed as:
\begin{equation}
	\label{5}
	\begin{aligned}
		I\left(X_1,X_2;O^l_{ADD}[t]\right)&=H\left(O^l_{ADD}[t]\right)-H\left(O^l_{ADD}[t] \mid X_1,X_2\right),\\
		I\left(X_1,X_2;O^l_{OR}[t]\right)&=H\left(O^l_{OR}[t]\right)-H\left(O^l_{OR}[t] \mid X_{1},X_2\right).
	\end{aligned}
\end{equation}
By applying the established inequality, we derive the following result:
\begin{equation}
	\label{6}
	I\left(X_1,X_2;O^l_{ADD}[t]\right) \ge I\left(X_1,X_2;O^l_{OR}[t]\right).
\end{equation}

The ADD operation preserves the joint information of input signals through current summation prior to spike generation, whereas the OR operation incurs greater information loss by independently processing inputs with logical disjunction. Therefore, from an information-theoretic perspective, the ADD operation demonstrates superior performance over the OR operation when implementing non-identity mappings with scaling transformations.

\subsection{Exclusive-OR Meta-Residuals}
\label{XOR}

SNNs rely on asynchronous binary spikes for transmitting and representing information. Since spikes are binary, additional computation and redundant spikes are unnecessary. However, the current residual structure used to construct deep SNNs doesn't account for this, completely following the residual extraction method of ResNet, resulting in some spike redundancy and redundant learning. To reduce spike redundancy and promote residual learning in the backbone branch, we propose selecting pre-learning residuals for the backbone branch via the XOR operation. Since these pre-learning residuals produce the desired residuals, we also refer to them as meta-residuals.

\begin{figure}[!htb]
	\centerline{\includegraphics[width=\textwidth]{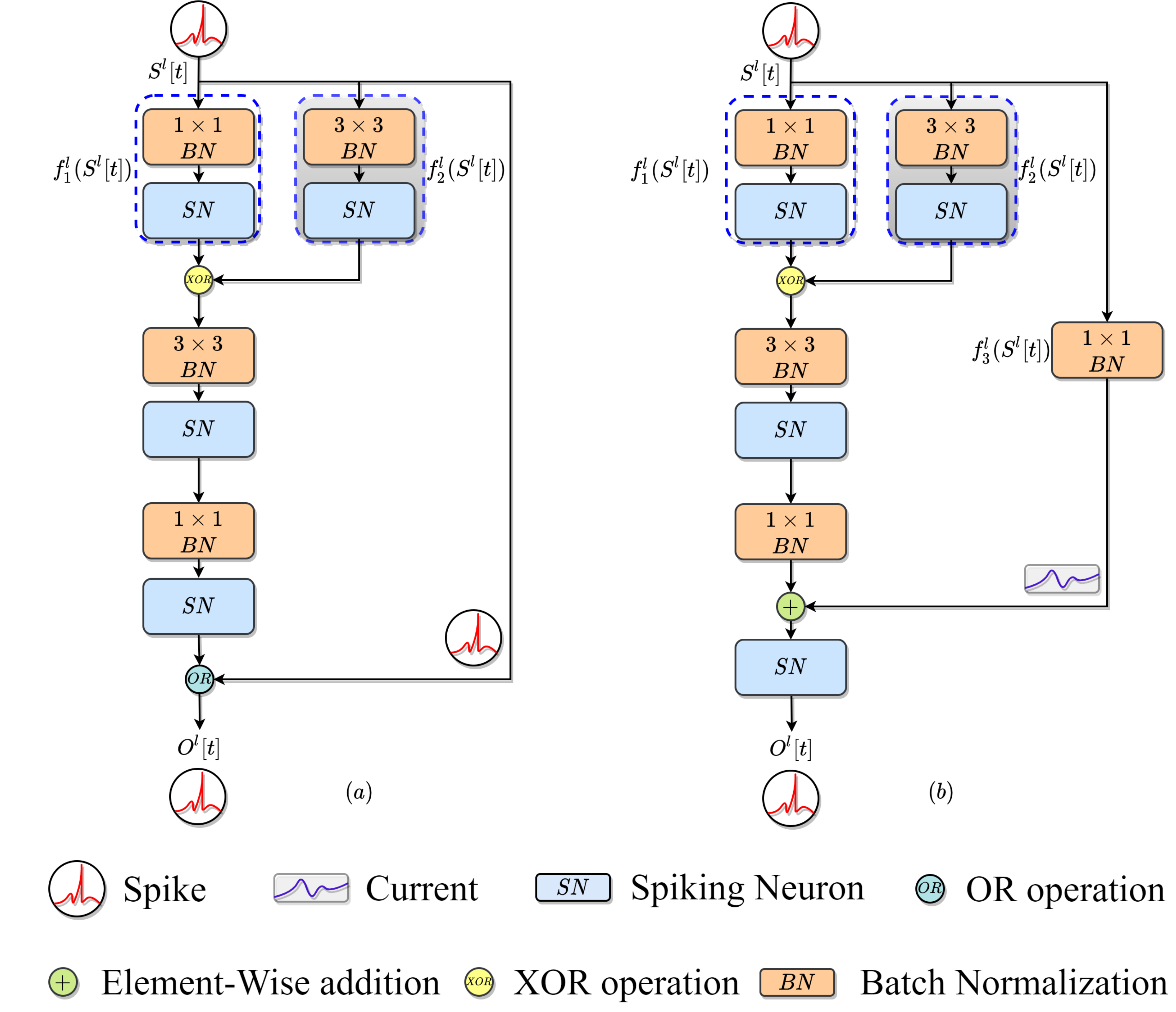}}
	\caption{Exclusive-OR meta-residuals structure. (a) OR shortcut connection. (b) ADD shortcut connection.}
	\label{fig4}
\end{figure}

The proposed XOR meta-residuals structure is shown in Fig.~\ref{fig4}. Since $f^{l}_{1}(S^{l}[t])$ is primarily used for channel transformation of feature maps with approximate input and output features, the output feature scale of $f^{l}_{2}(S^{l}[t])$ is the same as that of $f^{l}_{1}(S^{l}[t])$. The output features of the two branches are merged by the XOR operation to select the meta-residuals that need to be learned, which is expressed as:

\begin{equation}
	\label{equ3}
	\begin{aligned}
		MR&=f^{l}_{1}(S^{l}[t]) \oplus f^{l}_{2}(S^{l}[t]),\\
		\Longrightarrow  MR=f^{l}_{1}(S^{l}[t]) &+  f^{l}_{2}(S^{l}[t])-2( f^{l}_{1}(S^{l}[t]) \times f^{l}_{2}(S^{l}[t])).
	\end{aligned}
\end{equation}
where \textit{MR} represents the meta-residuals, providing meta-residuals for the backbone branch to specialize in learning residual information that is different from that of the shortcut branch, which both reduces spike redundancy and facilitates residual learning in the backbone branch.

Utilizing the OA shortcut connection method proposed in Section \ref{OR-ADD} for merging output spikes/currents, and the proposed XOR meta-residuals structure is shown in Fig.~\ref{fig4}. In the following, we combine these two shortcut connection methods to demonstrate that XOResNet, a deep SNN constructed based on the residual structure with XOR meta-residuals, does not suffer from the gradient vanishing/exploding problem during gradient-based training.

When the residual $\mathcal{F}^{l}(S^{l}[t])=0$ the identity mapping is completed by an OR connection, and the gradient of the output $O^{l}[t]$ of the \emph{l}-th residual block to the input $S^{l}[t]$ is computed as:
\begin{equation}
	\label{equ4}
	\begin{aligned}
		\frac{\partial O^{l}[t]}{\partial S^{l}[t]}&=\frac{\partial (\mathcal{F}^{l}(S^{l}[t]) +S^{l}[t] -  \mathcal{F}^{l}(S^{l}[t]) \times S^{l}[t])}{\partial S^{l}[t]}\\
		&=\frac{\partial S^{l}[t]}{\partial S^{l}[t]}=1
	\end{aligned}.
\end{equation}
Therefore, the OA shortcut connection can overcome the vanishing/exploding gradient problem, and in principle, XOResNet can be deepened to any desired depth.

\subsection{Network Architectures}
\begin{figure}[!ht]
	\centerline{\includegraphics[width=\textwidth]{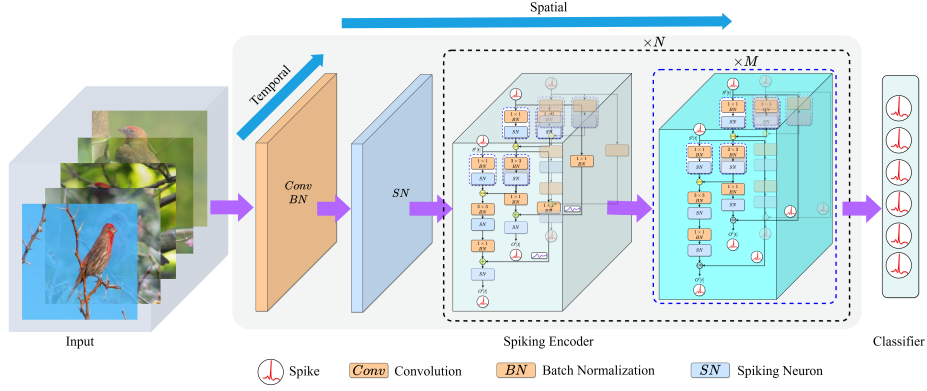}}
	\caption{The network structure of XOResNet and its unfolded formulation. $\times M$ denotes the number of iterations of the identity mapping, and $\times N$ denotes the number of downsample steps. Note that the network's parameters are shared at all time-steps.}
	\label{fig5}
\end{figure}

The proposed XOR meta-residuals block is used as the basic building block to stack the XOResNet to different depths, as shown in Table~\ref{tab2}, where $\times K$ denotes the number of iterative stackings. Images from different datasets have different sizes and corresponding output sizes. For instance, $28$ denotes an output size of $28\times28$, and other sizes follow a similar pattern. Downsampling is performed in Stage 2 and Stage 3 by a convolution operation with stride=2. In the stem stage, for an input image of $256\times256$ resolution, we first utilize a $4\times4$ convolution to reduce its size to $\frac{1}{4}$ of the original. The structure of the proposed XOResNet is illustrated in Fig. \ref{fig5}, unfolded in the temporal and spatial dimensions with shared network parameters in the temporal dimension. XOResNet comprises a spiking encoder network and a classifier network, where $\times N$ corresponds to ``Stage" in Table~\ref{tab2} and $\times M$ to $\times K$ in Table~\ref{tab2}.

\begin{table}[!htb]
	\Large
	\renewcommand{\arraystretch}{1.8}
	\centering
	\captionsetup{skip=2pt}
	\caption{The architecture of XOResNet with different depths.}
	\label{tab2}
	\resizebox{\textwidth}{!}
	{
		\begin{threeparttable}
			\begin{tabular}{c|c|c|c|c|c|c}
				\toprule
				Stage&Output size&11/20-layer&32/44-layer&50-layer&56-layer&110-layer\\
				\bottomrule
				Stem&28/32/56&\multicolumn{5}{c}{3$\times$3, 32, stride 2/4$\times$4, 64, stride 4}\\
				\hline
				\multirow{3}{*}{Stage 1}&\multirow{3}{*}{28/32/56}&\multirow{3}{*}{$\begin{bmatrix}
						1\times 1|3\times 3,&32 \\
						3\times 3, & 32\\
						1\times 1,&128
					\end{bmatrix}\times 1/2$}&\multirow{3}{*}{$\begin{bmatrix}
						1\times 1|3\times 3,&32 \\
						3\times 3, & 32\\
						1\times 1,&128
					\end{bmatrix}\times 3$}&\multirow{3}{*}{$\begin{bmatrix}
						1\times 1|3\times 3,&32 \\
						3\times 3, & 32\\
						1\times 1,&128
					\end{bmatrix}\times 3$}&\multirow{3}{*}{$\begin{bmatrix}
						1\times 1|3\times 3,&32 \\
						3\times 3, & 32\\
						1\times 1,&128
					\end{bmatrix}\times 3$}&\multirow{3}{*}{$\begin{bmatrix}
						1\times 1|3\times 3,&32 \\
						3\times 3, & 32\\
						1\times 1,&128
					\end{bmatrix}\times 19$}\\
				&&&&&&\\
				&&&&&&\\
				\hline
				\multirow{3}{*}{Stage 2}&\multirow{3}{*}{14/16/28}&\multirow{3}{*}{$\begin{bmatrix}
						1\times 1|3\times 3,&64 \\
						3\times 3, & 64\\
						1\times 1,&256
					\end{bmatrix}\times 1/2$}&\multirow{3}{*}{$\begin{bmatrix}
						1\times 1|3\times 3,&64 \\
						3\times 3, & 64\\
						1\times 1,&256
					\end{bmatrix}\times 4/8$}&\multirow{3}{*}{$\begin{bmatrix}
						1\times 1|3\times 3,&64 \\
						3\times 3, & 64\\
						1\times 1,&256
					\end{bmatrix}\times 10$}&\multirow{3}{*}{$\begin{bmatrix}
						1\times 1|3\times 3,&64 \\
						3\times 3, & 64\\
						1\times 1,&256
					\end{bmatrix}\times 12$}&\multirow{3}{*}{$\begin{bmatrix}
						1\times 1|3\times 3,&64 \\
						3\times 3, & 64\\
						1\times 1,&256
					\end{bmatrix}\times 14$}\\
				&&&&&&\\
				&&&&&&\\
				\hline
				\multirow{3}{*}{Stage 3}&\multirow{3}{*}{7/8/14}&\multirow{3}{*}{$\begin{bmatrix}
						1\times 1|3\times 3,&128 \\
						3\times 3, & 128\\
						1\times 1,&512
					\end{bmatrix}\times 1/2$}&\multirow{3}{*}{$\begin{bmatrix}
						1\times 1|3\times 3,&128 \\
						3\times 3, & 128\\
						1\times 1,&512
					\end{bmatrix}\times 3$}&\multirow{3}{*}{$\begin{bmatrix}
						1\times 1|3\times 3,&128 \\
						3\times 3, & 128\\
						1\times 1,&512
					\end{bmatrix}\times 3$}&\multirow{3}{*}{$\begin{bmatrix}
						1\times 1|3\times 3,&128\\
						3\times 3, & 128\\
						1\times 1,&512
					\end{bmatrix}\times 3$}&\multirow{3}{*}{$\begin{bmatrix}
						1\times 1|3\times 3,&128 \\
						3\times 3, & 128\\
						1\times 1,&512
					\end{bmatrix}\times 3$}\\
				&&&&&&\\
				&&&&&&\\
				\hline
				Classifier&1$\times$1&\multicolumn{5}{c}{average pool, 10/100-FC}\\
				\bottomrule
			\end{tabular}
			\begin{tablenotes}
				\Large
				\item The three output sizes respectively correspond to the Fashion-MNIST, CIFAR-10/100, and miniImageNet datasets.
			\end{tablenotes}
		\end{threeparttable}
	}
\end{table}

\section{Experiments and results}
\label{sec:Experiments and results}

We conduct extensive experiments on four publicly available datasets: Fashion-MNIST~\cite{40}, CIFAR-10~\cite{41}, CIFAR-100~\cite{41}, and miniImageNet~\cite{42}. For simplicity, we refer to the OR-connected ResNet as OR ResNet and the OA-connected ResNet as OA ResNet. Both architectures do not incorporate meta-residual configurations, and they differ solely in their non-identity mapping connections.

\subsection{Performance evaluation in classification tasks}

\begin{figure}[!htb]
	\centerline{\includegraphics[width=\textwidth]{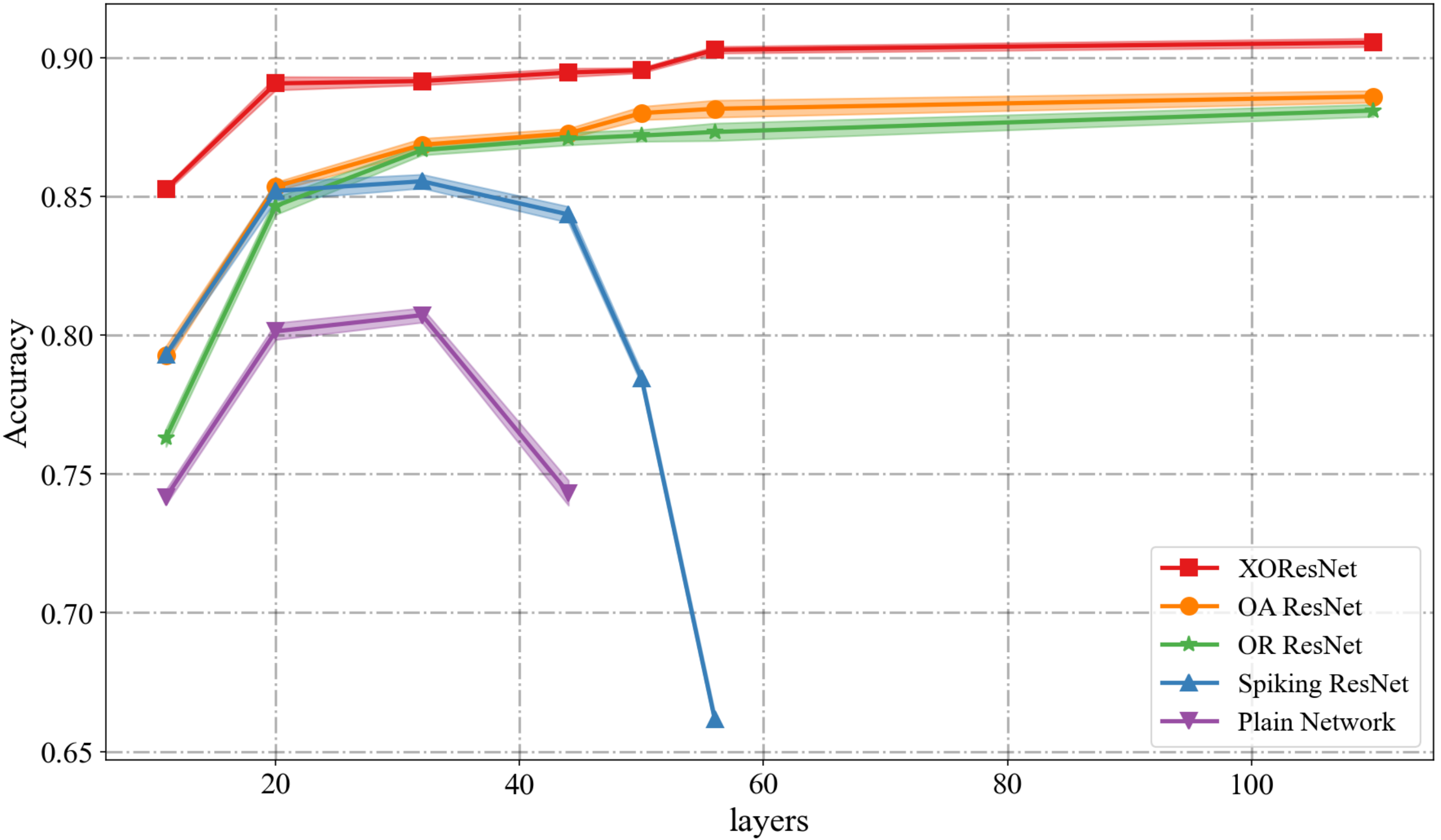}}
	\caption{Evaluation of models with varying depths on the CIFAR-10 dataset.}
	\label{fig6}
\end{figure}

\begin{figure}[!htbp]
	\begin{subfigure}[b]{\textwidth}
		\includegraphics[width=\textwidth]{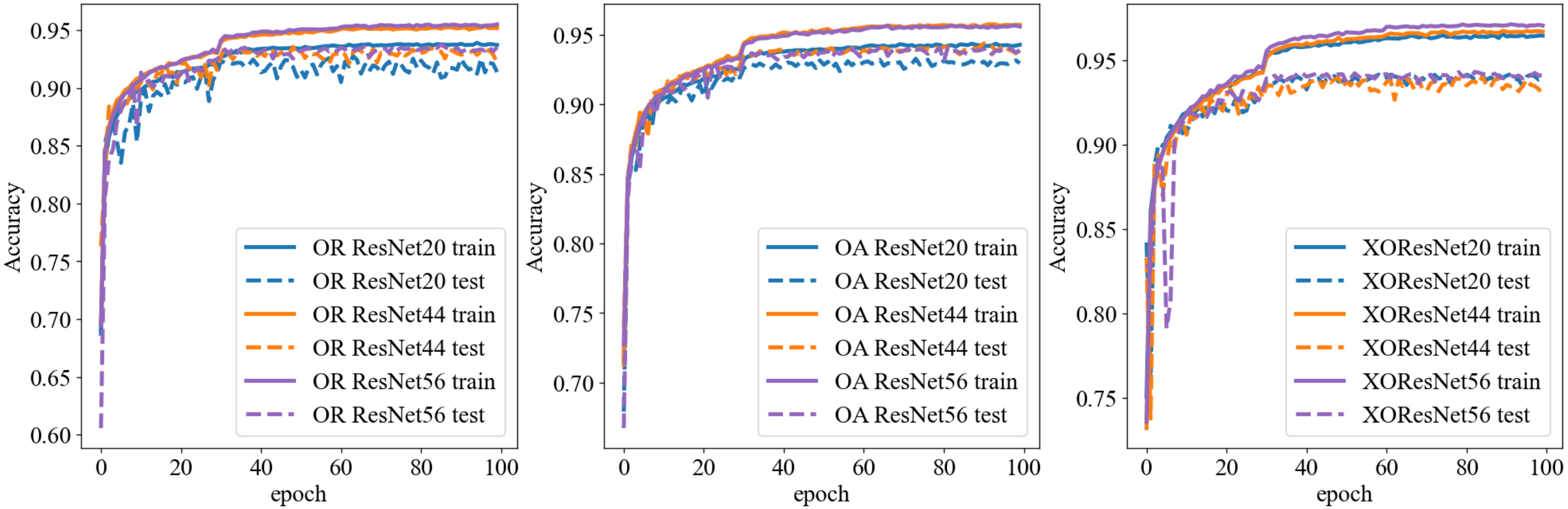}
		\caption{Fashion-MNIST}
		\label{a}
	\end{subfigure}
	\\
	\begin{subfigure}[b]{\textwidth}
		\includegraphics[width=\textwidth]{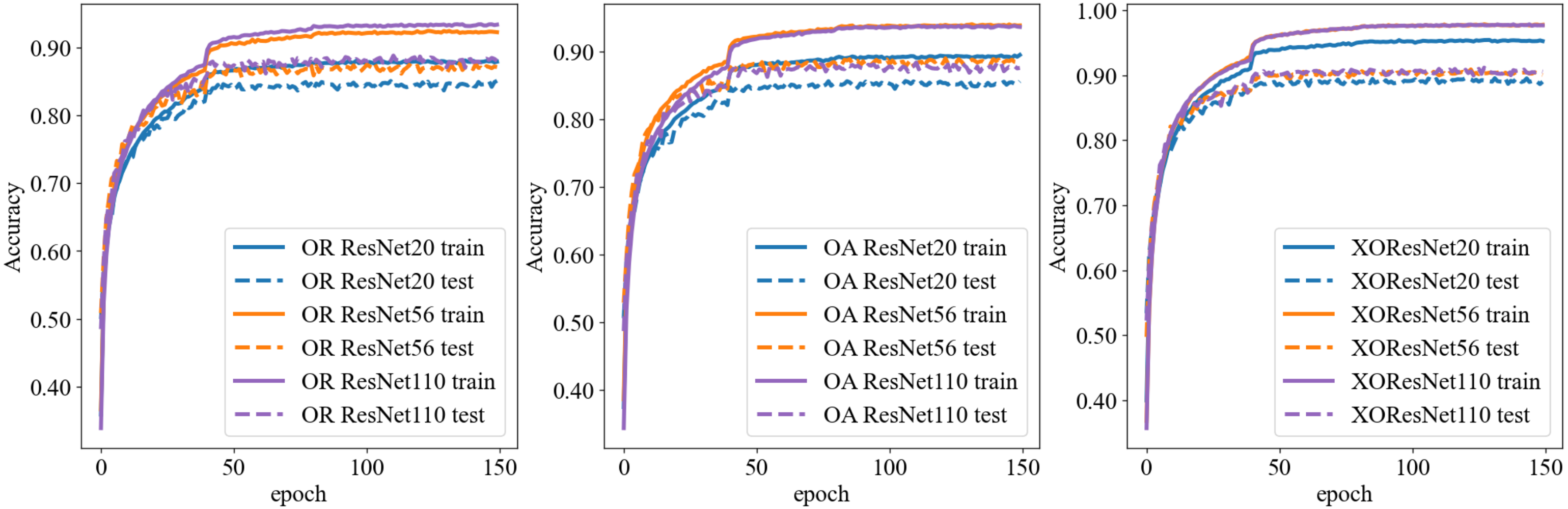}
		\caption{CIFAR-10}
		\label{b}
	\end{subfigure}
	\\
	\begin{subfigure}[b]{\textwidth}
		\includegraphics[width=\textwidth]{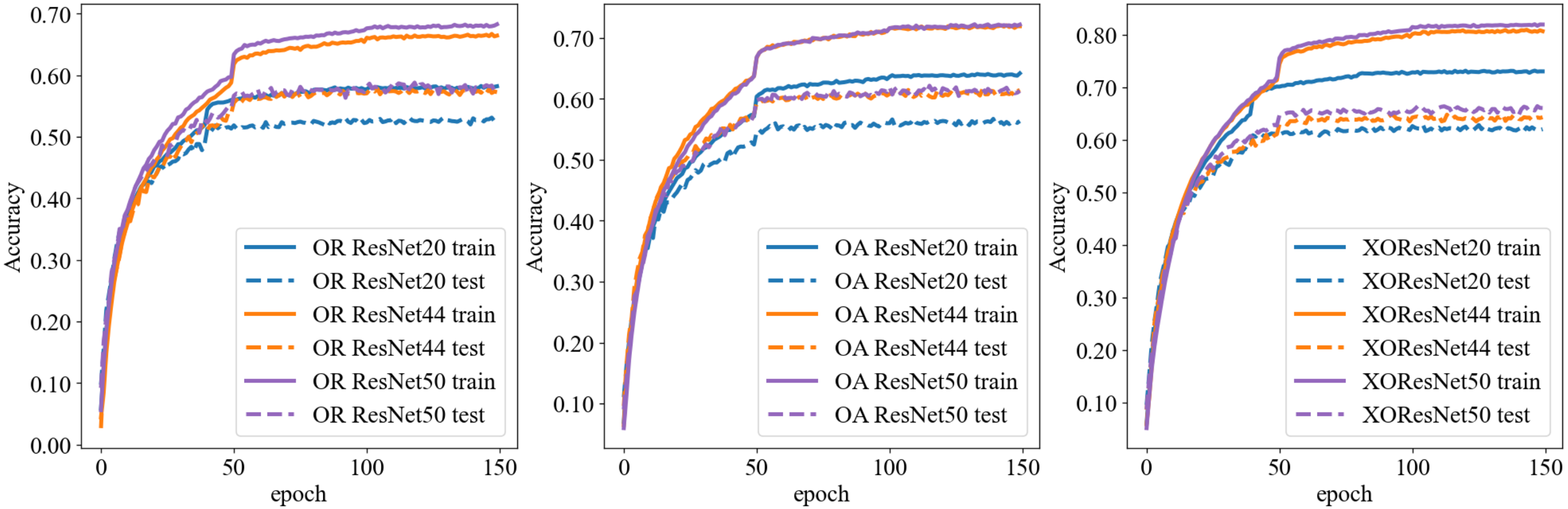}
		\caption{CIFAR-100}
		\label{c}
	\end{subfigure}
	\\
	\begin{subfigure}[b]{\textwidth}
		\includegraphics[width=\textwidth]{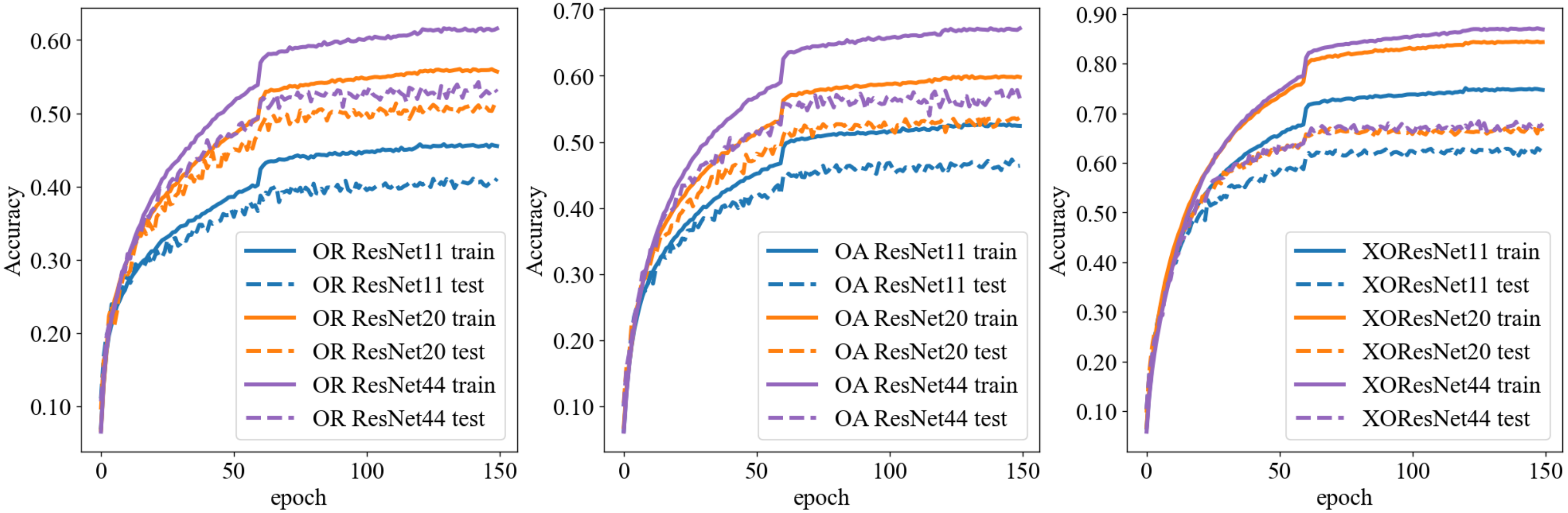}
		\caption{miniImageNet}
		\label{d}
	\end{subfigure}
	\caption{The training and test accuracy of different models with different depths on different datasets.}
	\label{fig7}
\end{figure}

We construct XOResNet models of varying depths according to the structure in Table~\ref{tab2}, and compare their performance with OA ResNet, OR ResNet, Spiking ResNet, and Plain Network models of the same depths, as shown in  Fig.~\ref{fig6}. We report the mean and standard deviation of model accuracy across 10 trials. Both Plain Network and Spiking ResNet suffer from performance degradation as model depth increases. However, OR ResNet, OA ResNet, and XOResNet do not exhibit this problem, as shown in Fig.~\ref{fig7}: the training and test accuracy of these models are not inferior to those of shallower models across the four datasets. This indicates that the OR and OA shortcut connections alleviate the vanishing/exploding gradient issues caused by deep network architectures. As shown in Fig.~\ref{fig6}, XOResNet significantly outperforms other models, indicating our proposed XOR meta-residuals structure facilitates deep SNN learning. Meanwhile, OA ResNet outperforms OR ResNet at the same depth, indicating OR shortcut solves the gradient problem but results in information loss for non-identity mapping, which the OA connection resolves.

\begin{table}[!ht]
	\Huge
	\renewcommand{\arraystretch}{1.4}
	\centering
	\captionsetup{skip=2pt}
	\caption{Performance comparison between the proposed method and previous works on different datasets.}
	\label{tab3}
	\resizebox{0.7\textwidth}{!}
	{
		\begin{threeparttable}
			\begin{tabular}{cccccccc}
				\toprule
				\textbf{Dataset}&\textbf{Method}&\textbf{Time-steps}&\textbf{Accuracy(\%)}\\
				\midrule
				\multirow{10}{*}{Fashion-MNIST}&Spiking ResNet\cite{51}&16&93.94\\
				&LISNN\cite{43}&20&92.07\\
				&ST-RSBP\cite{44} &1&90.13\\
				&SNN-BP\cite{45}&5&93.28\\
				&BackEISNN\cite{46}&30&93.04$\pm$0.31\\
				&TSSL-BP\cite{47}&5&92.69$\pm$0.09\\
				&OR-Spiking ResNet\cite{22}&16&94.21\\
				&OR ResNet50\cite{22}&8&93.88$\pm$0.02\\
				&\textbf{OA ResNet50(ours)}&8&94.11$\pm$0.06\\
				&\textbf{XOResNet50(ours)}&8&94.53$\pm$0.05\\
				\hline
				\multirow{11}{*}{CIFAR-10}&Spiking ResNet\cite{51}&16&88.65\\
				&ANN2SNN\cite{48}&30&82.95\\
				&ANN2SNN\cite{50}&400&77.43\\
				&ANN2SNN(ResNet20)\cite{16}&2000&87.46\\
				&NeuNorm\cite{49}&12&90.53\\
				&ANN2SNN(ResNet20)\cite{18}&2048&91.36\\
				&MS-ResNet56\cite{38}&6&90.4\\
				&OR-Spiking ResNet\cite{22}&16&89.72\\
				&OR ResNet110\cite{22}&8&88.09$\pm$0.22\\
				&\textbf{OA ResNet110(ours)}&8&88.60$\pm$0.21\\
				&\textbf{XOResNet110(ours)}&8&90.54$\pm$0.16\\
				\hline
				\multirow{6}{*}{CIFAR-100}&VGG11\cite{45}&12&63.97\\
				&ANN2SNN(ResNet20)\cite{18}&2048&67.82\\
				&MS-ResNet50\cite{38}&6&65.24\\
				&OR ResNet50\cite{22}&8&58.58$\pm$0.14\\
				&\textbf{OA ResNet50(ours)}&8&61.85$\pm$0.13\\
				&\textbf{XOResNet50(ours)}&8&66.30$\pm$0.13\\
				\hline
				\multirow{7}{*}{miniImageNet}&SNN-ResNet-12(5-way 1-shot )\cite{52}&16&48.37$\pm$0.24\\
				&SNN-ResNet-12(5-way 5-shot )\cite{52}&16&65.61$\pm$0.26\\
				&Matching networks\cite{42}&-&60.00\\
				&BASS\cite{53}&-&36.60$\pm$0.30\\
				&OR ResNet44\cite{22}&4&53.68$\pm$0.24\\
				&\textbf{OA ResNet44(ours)}&4&57.59$\pm$0.20\\
				&\textbf{XOResNet44(ours)}&4&68.00$\pm$0.21\\
				
				\bottomrule
			\end{tabular}
		\end{threeparttable}
		
	}
	
\end{table}

To evaluate the effectiveness of the proposed method for facilitating deep SNN learning, we compare it with existing state-of-the-art methods. Table~\ref{tab3} reports the results on the four datasets. On the Fashion-MNIST dataset, the highest recognition accuracy of our XOResNet50 is 94.53\%, which is far superior to other methods. On the CIFAR-10 dataset, our XOResNet achieves optimal recognition performance except for ANN2SNN proposed in \cite{18}. Notably, ANN2SNN in \cite{18} uses 2,048 time-steps, whereas our XOResNet uses only 8, reducing inference time to 1/256. \cite{16}, \cite{48} and \cite{50} also use ANN2SNN but with fewer time-steps and correspondingly weaker performance than that of literature \cite{18}, showing inference time-step is critical for ANN2SNN. This also indicates training deep SNNs with STBP is necessary. Similarly, on the 100-categorized CIFAR-100 dataset, our method is second only to the method presented in \cite{18}. The resolution of images in the aforementioned three datasets is low, and we need to evaluate XOResNet performance on a more complex dataset. Since ImageNet is too large, we use its subset miniImageNet as an alternative, containing 100 classes with 600 images each. We standardize image resolution to $256\times256$. The significant performance advantage of our XOResNet44 over methods based on few-shot learning shows that our model is effective at extracting features for both simple and complex images.

\begin{figure}[!htb]
	\centerline{\includegraphics[width=\textwidth]{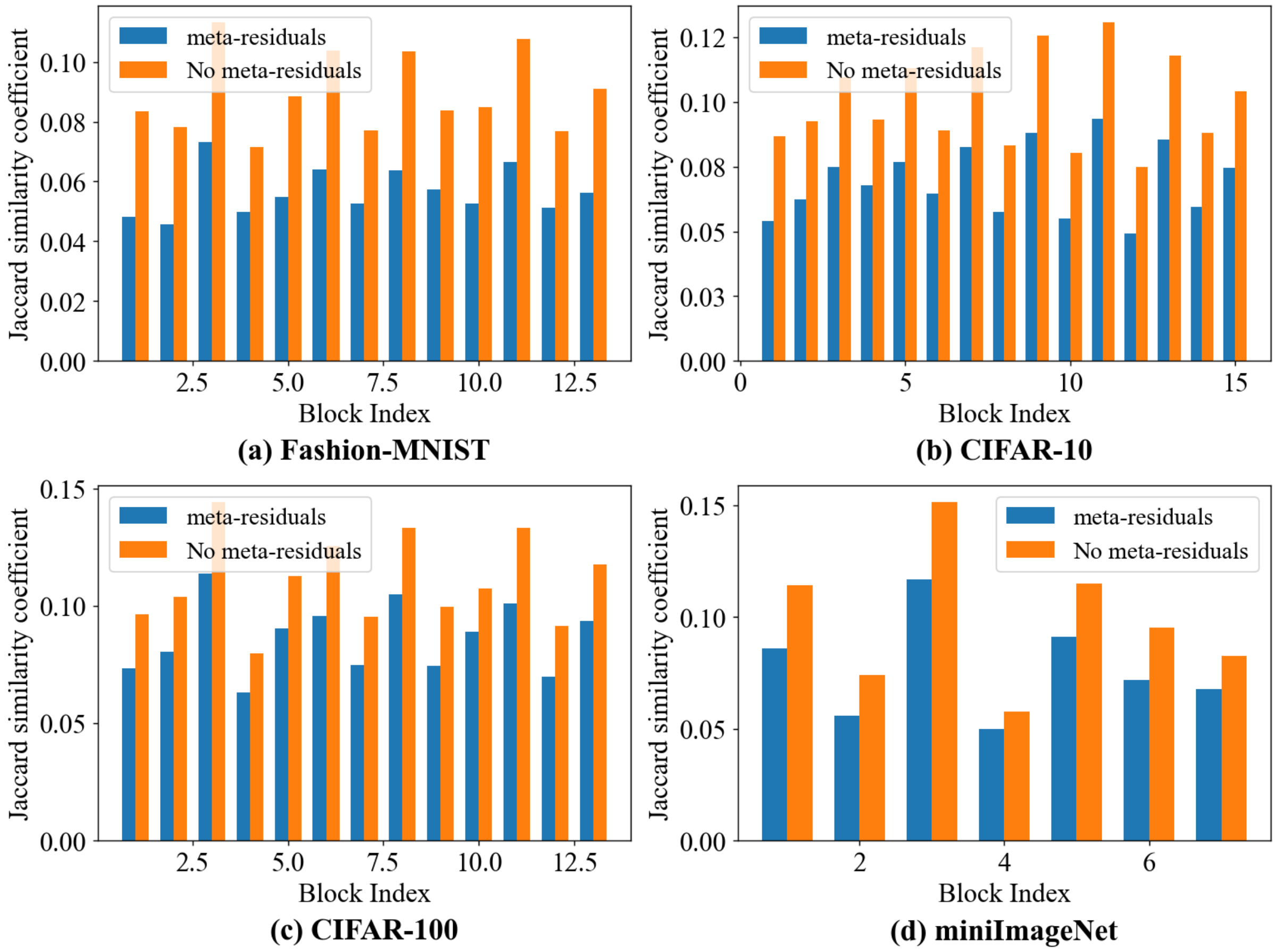}}
	\caption{The Jaccard similarity coefficient between the input and output binary spike features of the backbone branch in the residual block with identity mapping.}
	\label{fig8}
\end{figure}

To further demonstrate the effectiveness of the proposed XOR meta-residuals structure in reducing redundant learning, we compute the Jaccard similarity coefficient~\cite{54} between the input and output binary spike features of the backbone branch in the residual block with identity mapping. The Jaccard similarity coefficient is the intersection over union (IOU) of two binary spike features. A larger value indicates greater similarity, and vice versa. As shown in Fig.~\ref{fig8}, for the four datasets, the depths of the SNN models are 56, 50, 50, and 32, respectively, and the corresponding numbers of residual blocks with identity mapping are 15, 13, 13, and 7, respectively. On all four datasets, Jaccard similarity coefficients between input and output features of the backbone branch with XOR meta-residuals are smaller than those without meta-residuals, indicating XOR meta-residuals reduce redundancy learning of SNN and promote residual learning of the backbone branch.

\subsection{Ablation study}
The proposed XOResNet shows superior performance compared to other models and methods on four datasets: Fashion-MNIST, CIFAR-10, CIFAR-100, and miniImageNet. To futher validate the effectiveness of the proposed individual components, we perform ablation analysis on these four datasets.


\textbf{Fashion-MNIST} \quad The Fashion-MNIST dataset comprises grayscale images, each with a resolution of $28\times28$ pixels. Based on the structural setup in Table~\ref{tab2}, we construct five SNN networks with varying depths, as detailed in Table~\ref{tab4}. As a baseline method, the OR shortcut connection method solves the gradient problem. As demonstrated in Table~\ref{tab4}, SNNs employing OR shortcut connections for both identity and non-identity mappings exhibit no performance degradation when the network depth is increased from 11 to 50 layers. However, we demonstrate that the OR shortcut connection is inapplicable to non-identity mapping in Eq. (\ref{equ1}). We propose an OA connection method that addresses both identity and non-identity mappings. For identity mapping, the method merges the output spikes from two branches using OR shortcut connections, thereby eliminating spike redundancy while preserving the binary properties of the spikes. For non-identity mapping, the method aggregates the current sums from the two branches as inputs to the spiking neurons, effectively mitigating information loss. As shown in Table~\ref{tab4}, SNNs employing OA connections exhibit no performance degradation as the network depth increases from 11 to 50 layers, indicating that the proposed OA connection method effectively mitigates gradient-related issues. For identical network depths, SNNs with OA connections demonstrate superior recognition performance compared to those with OR connections, indicating that current merging proves more effective than spike merging for non-identity mappings. 

\begin{table}[!ht]
	\Huge
	\renewcommand{\arraystretch}{1.5}
	\centering
	\captionsetup{skip=2pt}
	\caption{The test accuracy of different depth models on the Fashion-MNIST dataset.}
	\label{tab4}
	\resizebox{\textwidth}{!}
	{
		\begin{threeparttable}
			\begin{tabular}{c||c|c||ccccc}
				\toprule
				\multirow{2}{*}{meta-residuals}&\multicolumn{2}{c||}{shortcut}&\multicolumn{5}{c}{depth}\\
				\cline{2-8}
				&im&nim&11&20&32&44&50\\
				\midrule
				\ding{55}&OR&OR &90.59$\pm$0.08\%&92.50$\pm$0.11\%&93.15$\pm$0.22\%&93.39$\pm$0.09\%&93.88$\pm$0.02\%\\		
				\ding{55}&OR&ADD&91.28$\pm$0.03\%&93.18$\pm$0.05\%&93.59$\pm$0.20\%&93.84$\pm$0.12\%&94.11$\pm$0.06\%\\			
				\ding{51}&OR&ADD&92.98$\pm$0.04\%&93.63$\pm$0.27\%&93.79$\pm$0.20\%&93.92$\pm$0.07\%&94.53$\pm$0.05\%\\
				\bottomrule
			\end{tabular}
		\end{threeparttable}
	}
\end{table}

\begin{figure}[!ht]
	\centerline{\includegraphics[width=\textwidth]{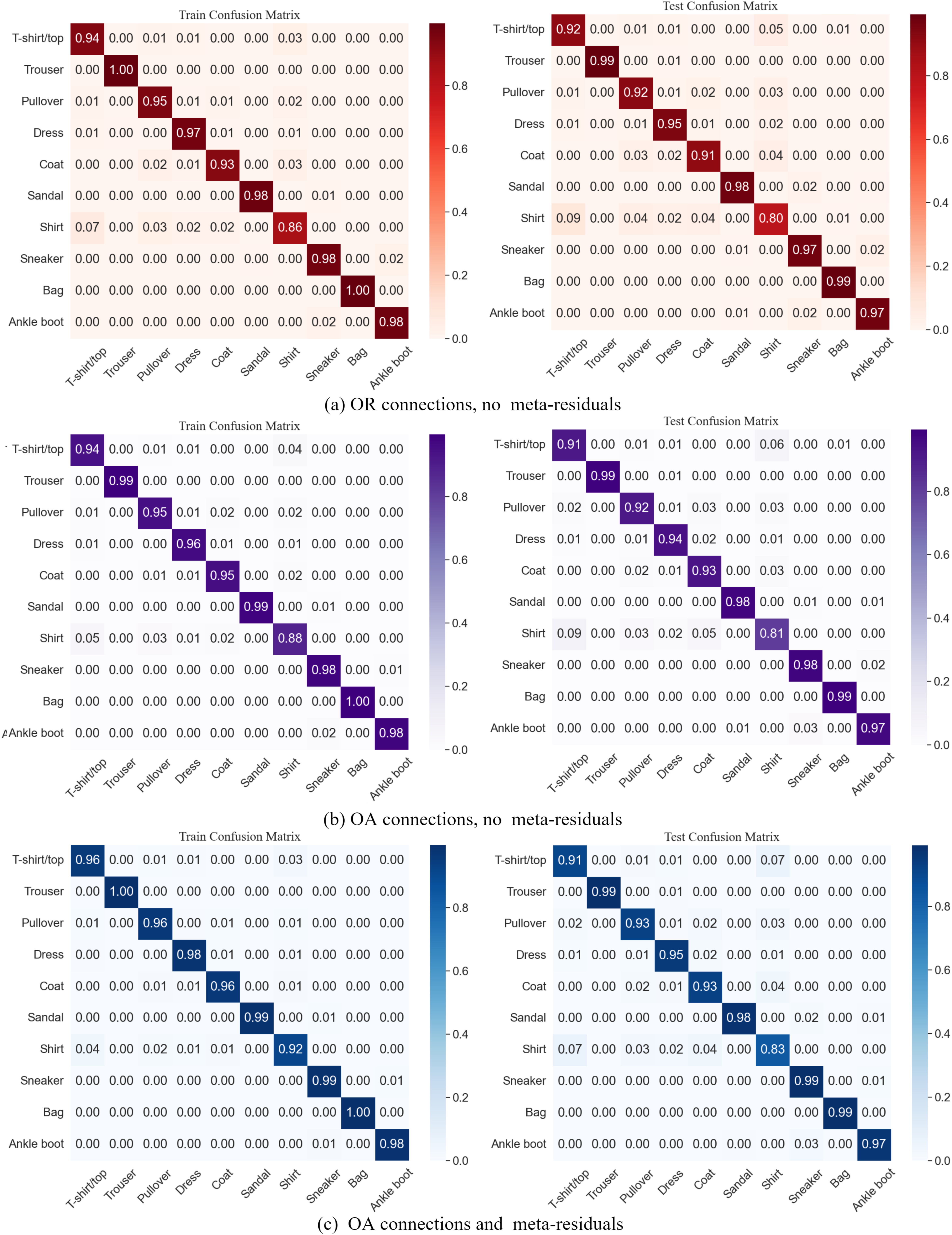}}
	\caption{Confusion matrices of 50-layer networks on the Fashion-MNIST dataset. The left column represents the training dataset, and the right column represents the test dataset.}
	\label{fig9}
\end{figure}

We further integrate the meta-residual and OA connections into the SNN. As shown in Table~\ref{tab4}, for SNNs with identical depths, networks incorporating meta-residuals markedly outperform those without meta-residuals in terms of recognition performance. This indicates that the proposed meta-residual structure promotes learning in deep SNNs. Furthermore, Fig.~\ref{fig9} presents the confusion matrices of 50-layer networks under different configurations. The three network configurations achieved average test accuracies of 94.00\%, 94.20\%, and 94.70\% on the 10-class classification task, further validating the effectiveness of the proposed components.

\begin{table}[!ht]
	\Huge
	\renewcommand{\arraystretch}{1.5}
	\centering
	\captionsetup{skip=2pt}
	\caption{The test accuracy of different depth models on the CIFAR-10 dataset.}
	\label{tab5}
	\resizebox{\textwidth}{!}
	{
		\begin{threeparttable}
			\begin{tabular}{c||c|c||ccccccc}
				\toprule
				\multirow{2}{*}{meta-residuals}&\multicolumn{2}{c||}{shortcut}&\multicolumn{7}{c}{depth}\\
				\cline{2-10}
				&im&nim&11&20&32&44&50&56&110\\
				\midrule
				\ding{55}&OR&OR&76.31$\pm$0.32\%&84.65$\pm$0.31\%&86.67$\pm$0.18\%&87.07$\pm$0.23\%&87.19$\pm$0.21\%&87.32$\pm$0.32\%&88.09$\pm$0.22\%\\
				\ding{55}&OR&ADD&79.28$\pm$0.33\%&85.36$\pm$0.16\%&86.86$\pm$0.23\%&87.27$\pm$0.17\%&88.00$\pm$0.24\%&88.16$\pm$0.31\%&88.60$\pm$0.21\%\\
				\ding{51}&OR&ADD&85.26$\pm$0.11\%&89.07$\pm$0.24\%&89.16$\pm$0.14\%&89.46$\pm$0.15\%&89.54$\pm$0.09\%&90.28$\pm$0.12\%&90.54$\pm$0.16\%\\
				\bottomrule
			\end{tabular}
		\end{threeparttable}
	}
\end{table}

\newpage

\begin{figure}[!ht]
	\centerline{\includegraphics[width=\textwidth]{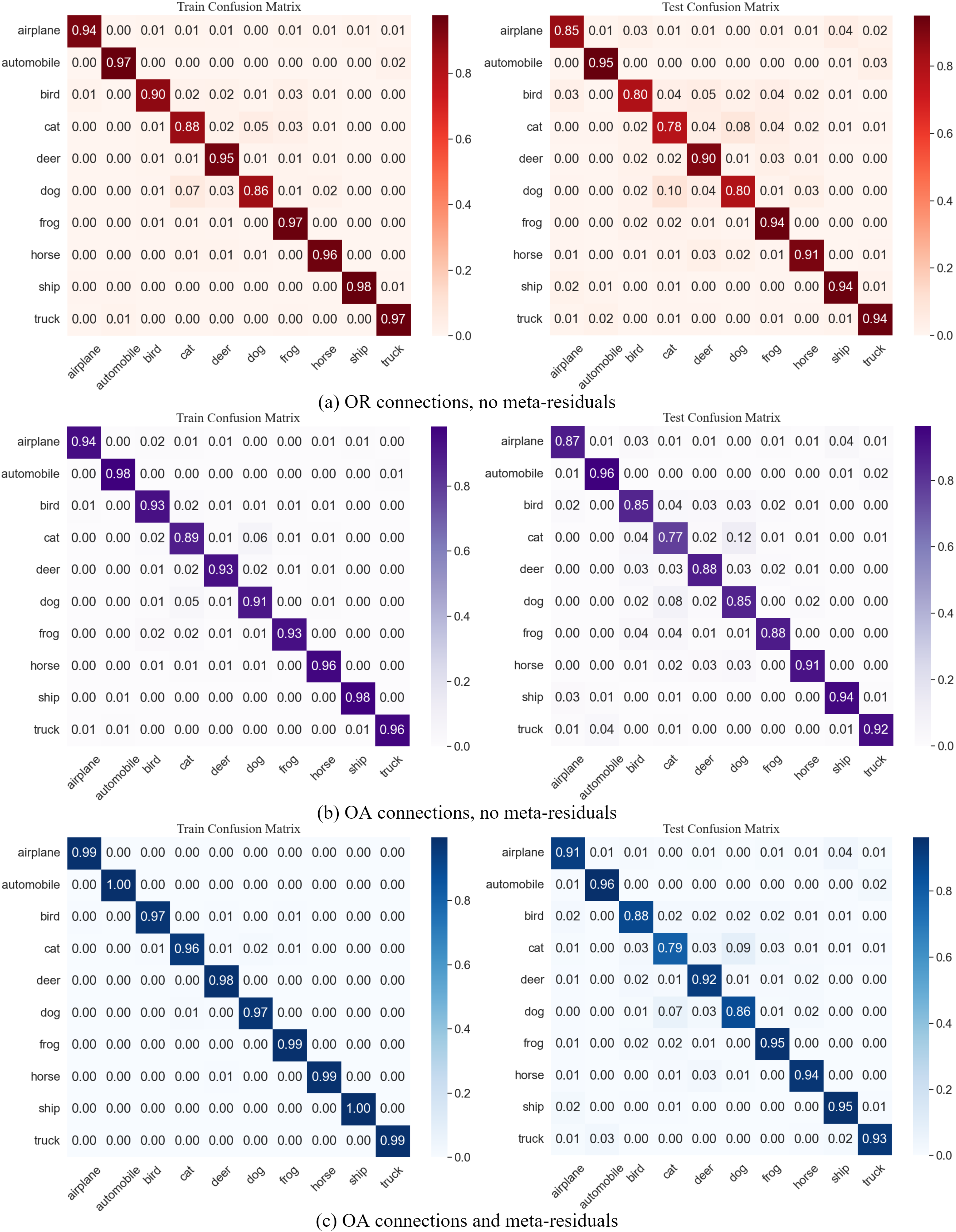}}
	\caption{Confusion matrices of 110-layer networks on the CIFAR-10 dataset. The left column represents the training dataset, and the right column represents the test dataset.}
	\label{fig10}
\end{figure}

\textbf{CIFAR-10} \quad We construct seven networks with varying depths, as shown in Table~\ref{tab5}, using different components based on the structural configuration in Table~\ref{tab2}. The networks exhibit no performance degradation as the model depth increases from 11 to 110 layers, indicating that both OR shortcut connections and OA shortcut connections can effectively address the gradient issues in deep models. For networks of identical depth, those using the OA connection method consistently demonstrate superior recognition performance compared to networks employing the OR connection method. This indicates that current merging proves more effective than spike merging for non-identity mapping connections. The introduction of the meta-residual component further improves the recognition performance of networks at each depth level, demonstrating that the proposed meta-residual component enhances residual learning in deep SNNs. Fig.~\ref{fig10} shows the confusion matrices of three 110-layer networks with different components. The average recognition accuracies of these networks on the CIFAR-10 classification task are 88.10\%, 88.30\%, and 90.90\%, respectively, which again demonstrates the effectiveness of our proposed individual components.

\begin{figure}[!htb]
	\centerline{\includegraphics[width=\textwidth]{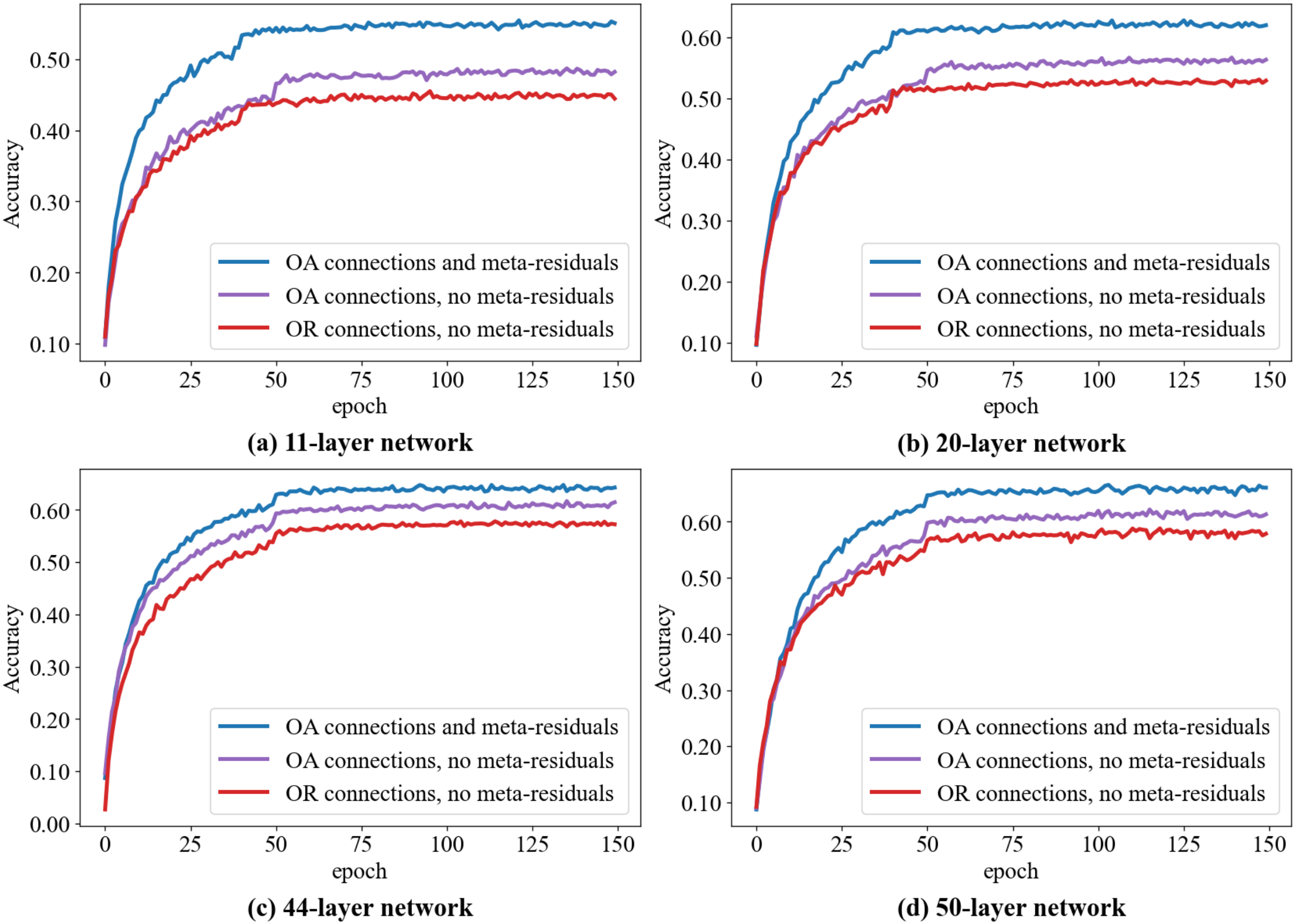}}
	\caption{Evaluation of models with varying depths on the CIFAR-100 dataset.}
	\label{fig11}
\end{figure}

\begin{table}[!htb]
	\Huge
	\renewcommand{\arraystretch}{1.5}
	\centering
	\captionsetup{skip=2pt}
	\caption{The test accuracy of different depth models on the CIFAR-100 dataset.}
	\label{tab6}
	\resizebox{\textwidth}{!}
	{
		\begin{threeparttable}
			\begin{tabular}{c||c|c||ccccc}
				\toprule
				\multirow{2}{*}{meta-residuals}&\multicolumn{2}{c||}{shortcut}&\multicolumn{5}{c}{depth}\\
				\cline{2-8}
				&im&nim&11&20&32&44&50\\
				\midrule
				\ding{55}&OR&OR&45.19$\pm$0.11\%&53.02$\pm$0.11\%&54.74$\pm$0.09\%&57.69$\pm$0.09\%&58.58$\pm$0.14\%\\
				\ding{55}&OR&ADD&48.51$\pm$0.11\%&56.47$\pm$0.14\%&59.14$\pm$0.07\%&61.36$\pm$0.15\%&61.85$\pm$0.13\%\\
				\ding{51}&OR&ADD&55.17$\pm$0.11\%&62.07$\pm$0.22\%&62.56$\pm$0.07\%&64.62$\pm$0.12\%&66.30$\pm$0.13\%\\
				\bottomrule
			\end{tabular}
		\end{threeparttable}
	}
\end{table}

\textbf{CIFAR-100} \quad Compared to the CIFAR-10 dataset with 10 classes, the CIFAR-100 dataset has 100 classes but only 600 samples per class, making model training and learning more challenging. As illustrated in Fig.~\ref{fig11}, across four networks with varying depths, the OA connection method consistently outperforms the OR method in model performance throughout all learning stages, while the introduction of the meta-residual component markedly enhances learning capabilities. Table~\ref{tab6} further demonstrates that OA shortcut connections outperform OR shortcut connections and that the meta-residual component enhances residual learning in the model, which aligns with the conclusions drawn from Table~\ref{tab4} and Table~\ref{tab5}.

\begin{figure}[!htb]
	\centerline{\includegraphics[width=\textwidth]{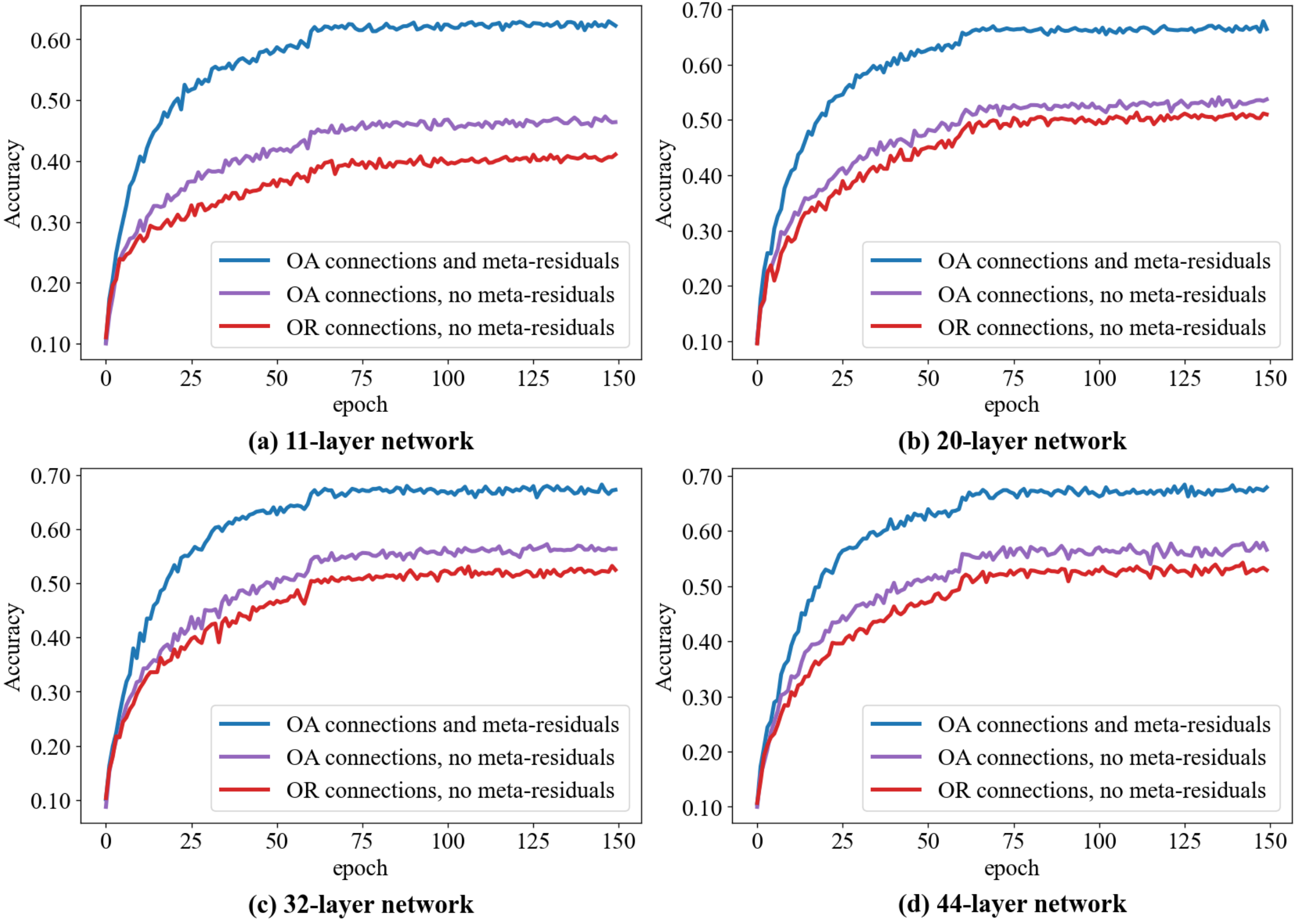}}
	\caption{Evaluation of models with varying depths on the miniImageNet dataset.}
	\label{fig12}
\end{figure}

\textbf{miniImageNet} \quad The image resolution for the aforementioned three datasets does not exceed $32\times32$. We evaluate model performance on the more complex miniImageNet dataset, standardizing image resolution to $256\times256$. On higher-resolution and more complex images, as demonstrated in Table~\ref{tab7}, the advantages of the OA connection method and the meta-residual component become even more pronounced. This superiority persists throughout the entire learning process of the model, as illustrated in Fig.~\ref{fig12}.

\begin{table}[!htb]
	\Huge
	\renewcommand{\arraystretch}{1.5}
	\centering
	\captionsetup{skip=2pt}
	\caption{The test accuracy of different depth models on the miniImageNet dataset.}
	\label{tab7}
	\resizebox{0.8\textwidth}{!}
	{
		\begin{threeparttable}
			\begin{tabular}{c||c|c||cccc}
				\toprule
				\multirow{2}{*}{meta-residuals}&\multicolumn{2}{c||}{shortcut}&\multicolumn{4}{c}{depth}\\
				\cline{2-7}
				&im&nim&11&20&32&44\\
				\midrule
				\ding{55}&OR&OR&40.86$\pm$0.16\%&51.04$\pm$0.14\%&52.71$\pm$0.25\%&53.68$\pm$0.24\%\\
				\ding{55}&OR&ADD&46.95$\pm$0.18\%&53.61$\pm$0.19\%&56.81$\pm$0.20\%&57.59$\pm$0.20\%\\
				\ding{51}&OR&ADD&62.83$\pm$0.06\%&67.03$\pm$0.28\%&67.89$\pm$0.14\%&68.00$\pm$0.21\%\\
				\bottomrule
			\end{tabular}
		\end{threeparttable}
	}
\end{table}

Through extensive experiments, we demonstrate that the proposed XOR meta-residuals structure can promote the learning of deep SNNs. Meanwhile, systematic ablation studies reveal the effectiveness of our proposed components.

\section{Discussion}
\label{sec:discussion}

The structure and ideas of ResNet in ANNs are inspirational and referential for constructing deep SNNs. 
Nevertheless, disregarding the spike binary property and spatio-temporal dynamics, Spiking ResNet, which exactly mimics the connection structure of ResNet (Fig.~\ref{fig1}), still suffers from performance degradation problem (Fig.~\ref{fig2}). The OR operation (Fig.~\ref{fig3}(a)) for solving performance degradation and maintaining the spike binary property is effective but causes information loss in the non-identity mapping (Eq.~(\ref{equ1}) and Eq.~(\ref{6})). We propose the OR-ADD (OA) connection method for shortcut connections in deep SNNs, as depicted in Eq.~(\ref{equ2}) and Fig.~\ref{fig3}(b)(c): (1) For shortcut connection with identity mapping, the output spikes from the two branches are merged using an OR operation, maintaining the binary spike property and avoiding redundancy. (2) For non-identity mapping connections requiring dimensional transformation, the sum of both branches' output currents serves as input to spiking neurons, avoiding information loss. We conduct a comprehensive analysis to elucidate the effectiveness of the proposed OA connection method in solving the vexing vanishing/exploding gradient problem. The innovative approach, as succinctly outlined in Eq.~(\ref{equ4}), offers a promising solution for constructing deep SNNs.

For residual learning of the backbone branch, given the consideration of binary spike communication, our innovative proposal is to select residual features via the XOR operation, an approach aimed at averting excessive computation and reducing spike redundancy. We introduce the concept of meta-residuals, as denoted in Eq.~(\ref{equ3}), which refers to the selected residual features that are yet to be learned. The integration of the proposed OA shortcut and meta-residuals concepts within the residual block leads to the formulation of the XOR meta-residuals structure, as visually represented in Fig.~\ref{fig4}. Employing this novel structure, we can construct XOResNet models of varying depths, the specifics of which are detailed in Table~\ref{tab2} and Fig. \ref{fig5}.

We conduct extensive experiments on four datasets: Fashion-MNIST, CIFAR-10, CIFAR-100, and miniImageNet. In comparison to state-of-the-art methods, as shown in Table~\ref{tab3}, our XOResNet method outperforms state-of-the-art deep SNNs trained with STBP on all four datasets. Meanwhile, Fig.~\ref{fig6} shows that both Plain Network and Spiking ResNet suffer from performance degradation, which can be solved by both OR connection and OA connection. The OA connection further compensates for the information loss problem in the OR connection. We rigorously monitor the model's performance changes during training. As shown in Fig.~\ref{fig7}, the deep model's performance is not inferior to the shallow model at any stage, matching exactly with the results and conclusions in Fig.~\ref{fig6}. We further demonstrate the contribution of the XOR meta-residuals structure to facilitating backbone branch residual learning via Jaccard similarity coefficients (Fig.~\ref{fig8}). Ablation experiments (Figs.~\ref{fig9}-\ref{fig12}, Tables.~\ref{tab4}-\ref{tab7}) demonstrate the effectiveness of the proposed OA in addressing degradation and information loss, and the contribution of XOR meta-residuals for residual learning. Our method provides an empirically supported reference for constructing deep SNNs, preserving the binary nature of spikes while mitigating information loss, spike redundancy, and redundant learning.

\section{Conclusion}
\label{sec:conclusion}
Spiking neural networks are considered promising models for achieving machine intelligence. The residual connection and design of ResNet in ANNs provide valuable insights for constructing deep SNNs using gradient-based training. In this work, we consider the specificity of spike communication and propose the OA shortcut connection method for SNNs. This maintains the binary property of spikes without causing redundancy or information loss. We also propose the XOR meta-residuals to facilitate residual learning in deep SNNs by selecting pre-learning features. Integrating these ideas, we propose the XOR meta-residuals structure and use it to construct deep XOResNet. Extensive experiments on four datasets demonstrate the superiority and efficiency of XOResNet, which can be deepened in principle to arbitrary depth. For future work, we will use XOResNet for research on biologically plausible Few-Shot Learning algorithms.

\section*{CRediT authorship contribution statement}
\textbf{Jianfang~Wu:} Writing - original draft, Software, Methodology, Formal analysis, Visualization, Validation, Conceptualization.
 \textbf{Junsong~Wang:} Formal analysis, Writing - review \& editing, Validation, Supervision,  Funding acquisition, Project administration.

\section*{Declaration of competing interest}
The authors declare that they have no known competing financial interests or personal relationships that could have appeared to influence the work reported in this paper.

\section*{Data availability}
The data used in this study are publicly available.

\section*{Acknowledgments}
This work was supported by Shenzhen Natural Science Foundation Project (Grant No. JCYJ20250604145123031), Shenzhen Science and Technology Major Project (Grant No. KJZD20230923114615032), the Major Project of Science and Technology Research Program of  Chongqing Education Commission (Grant No. KJZDM202302001), the National Natural Science Foundation of China (Grant No. 61876132), the Shenzhen University of Technology Self-made Experimental Instruments and Equipment Project (Grant No. JSZZ202301006), the Open Fund of National Engineering Laboratory for Big Data System Computing Technology (Grant No. SZU-BDSC-OF2024-13), and the Guangxi Key Laboratory of Brain-inspired Computing and Intelligent Chips (Grant No. BCIC-24-K8).

%
%


\bibliography{refs}

\end{document}